%% file: main.tex
\documentclass[10pt,twocolumn,letterpaper]{article}

\usepackage[T1]{fontenc} 

\usepackage[pagenumbers]{cvpr}   

\usepackage{mathtools}
\usepackage{multirow}
\usepackage{colortbl}
\usepackage{microtype}
\usepackage{dsfont}
\usepackage{tabularx}
\usepackage{tcolorbox}
\tcbuselibrary{listings,breakable}

\setlist[itemize]{noitemsep,leftmargin=*,topsep=0em}

\newcolumntype{C}{>{\centering\arraybackslash}X}

\definecolor{cprop}{RGB}{232,240,254}    
\definecolor{coffline}{RGB}{255,246,229} 
\definecolor{conline}{RGB}{234,247,237}  

\definecolor{q3orange}{HTML}{E8842C}
\definecolor{q25blue}{HTML}{4E9BB9}
\definecolor{pinZS}{HTML}{EB6834}
\definecolor{pinS}{HTML}{4A3AA7}
\definecolor{pinR}{HTML}{C0392B}
\definecolor{glancegreen}{HTML}{2FA35C}
\definecolor{lookbackorange}{HTML}{E8951F}
\definecolor{retrievalpurple}{HTML}{8B7FD7}
\definecolor{epsbg}{HTML}{FDE8E8}
\definecolor{epsframe}{HTML}{F0AFAF}
\definecolor{epsink}{HTML}{D6336C}

\newcommand{\legendbox}[2]{%
  {\setlength{\fboxsep}{1.5pt}%
   \raisebox{0pt}[0pt][0pt]{\colorbox{#1}{#2}}}%
}
\newcommand{\epsbadge}{{\setlength{\fboxsep}{1.5pt}\setlength{\fboxrule}{0.6pt}%
  \fcolorbox{epsframe}{epsbg}{\textcolor{epsink}{$\varepsilon$}}}}

\newtcblisting{promptbox}[1]{listing only, breakable, colback=gray!4, colframe=gray!55,
  title={#1}, fonttitle=\bfseries\small, left=1mm, right=1mm,
  listing options={basicstyle=\ttfamily\scriptsize, breaklines=true, columns=fullflexible}}

\definecolor{cvprblue}{rgb}{0.21,0.49,0.74}
\usepackage[pagebackref,breaklinks,colorlinks,allcolors=cvprblue]{hyperref}

\def\confName{CVPR}
\def\confYear{2026}

\title{\texorpdfstring{\makebox[\textwidth][c]{StreamScout: Learning When to Look Deeper for Streaming Video Understanding}}{StreamScout: Learning When to Look Deeper for Streaming Video Understanding}}

\author{
    Ce Zhang\textsuperscript{1,*}\quad
    Jing Bi\textsuperscript{2}\quad
    Jinxi He\textsuperscript{1}\quad
    Jianshu Zhang\textsuperscript{3}\quad
    Jingyang Lin\textsuperscript{2}\quad
    Yunzhong Xiao\textsuperscript{1}\\
    Minghao Fu\textsuperscript{4}\quad
    Yaqi Xie\textsuperscript{1}\quad
    Zhentao Xie\textsuperscript{5}\quad
    Weicong Chen\textsuperscript{5}\quad
    Katia Sycara\textsuperscript{1,$\dagger$}\quad
    Ming Zhou\textsuperscript{5,$\dagger$}\\[5pt]
    \textsuperscript{1}Carnegie Mellon University\quad
    \textsuperscript{2}University of Rochester\quad
    \textsuperscript{3}Northwestern University\\
    \textsuperscript{4}University of California, San Diego\quad
    \textsuperscript{5}TikTok
}

\begin{document}
\maketitle
{\let\thefootnote\relax
\footnotetext{\textsuperscript{*}Work done in part during an internship at TikTok.}
\footnotetext{\textsuperscript{$\dagger$}Equal advising.}}

\input{secs/0_abstract}

\input{secs/1_introduction}
\input{secs/2_related}

\input{secs/3_method}

\input{secs/4_experiments}
\input{secs/5_conclusion}

{
    \small
    \bibliographystyle{ieeenat_fullname}
    \bibliography{main}
}

\clearpage
\input{secs/X_appendix}

\end{document}

%% file: secs/0_abstract.tex
\begin{abstract}
Streaming video understanding requires answering questions that arrive at arbitrary moments over an unbounded video stream. Existing systems primarily focus on what to retain in a bounded memory, yet access that memory using the same fixed-cost procedure for every query, despite substantial variation in the evidence required. We argue that deciding how deeply to access memory for each query is as important as deciding what the memory should store. To this end, we introduce StreamScout, an adaptive inference framework that maintains only a lightweight textual timeline in context as the stream unfolds. At query time, StreamScout progressively augments the timeline with up to three increasingly informative visual views: a glance at recent frames, a uniform look-back over the past stream, and query-salient retrieval. At each stage, the model answers immediately if the available evidence is sufficient; otherwise, it escalates to the next view. To improve this stop-or-escalate policy, we probe the cascade on an auxiliary set and distill the model's empirical competence boundary into supervision for a lightweight LoRA adaptation, yielding StreamScout-S. We further refine the policy through reinforcement learning, allowing the model to explore stopping behaviors beyond imitation of the distilled decisions, yielding StreamScout-R. 
Across three backbones and three streaming benchmarks, StreamScout and its variants consistently outperform prior streaming methods while substantially reducing inference cost and token consumption; on OVO-Bench, for instance, StreamScout-S improves Qwen3-VL-8B by 14.65 points while using 59\% fewer tokens than uniform sampling and answering in 1.04\,s on average.
\end{abstract}

%% file: secs/1_introduction.tex
\section{Introduction}
\label{sec:intro}

Multi-modal large language models (MLLMs)~\cite{bai2025qwen3,bai2025qwen2,li2025llavaonevision,chen2024internvl} have extended visual understanding from static images to video, and streaming video is now emerging as a first-class input: a model watches an unbounded stream, such as footage from a wearable camera, a live broadcast, or a monitoring feed, and must answer questions that arrive at arbitrary moments~\cite{chen2024videollm,qian2025dispider,lin2026streamingbench,niu2025ovo}. The setting imposes two fundamental constraints. First, the stream cannot be retained in full: even at a modest rate of 0.5 fps, one hour produces 1,800 frames, and the stream may continue indefinitely. Second, questions are not known in advance, so the system cannot determine beforehand which moments or details will later become relevant. 
A practical streaming system must therefore continuously compress an unbounded input stream into bounded memory, while deciding what information to retain as the stream unfolds and what evidence to inspect when a question arrives.
A streaming system must decide both what information to retain as the stream unfolds and what evidence to inspect when a question arrives.

\begin{figure}[t]
    \centering
    \includegraphics[width=\linewidth]{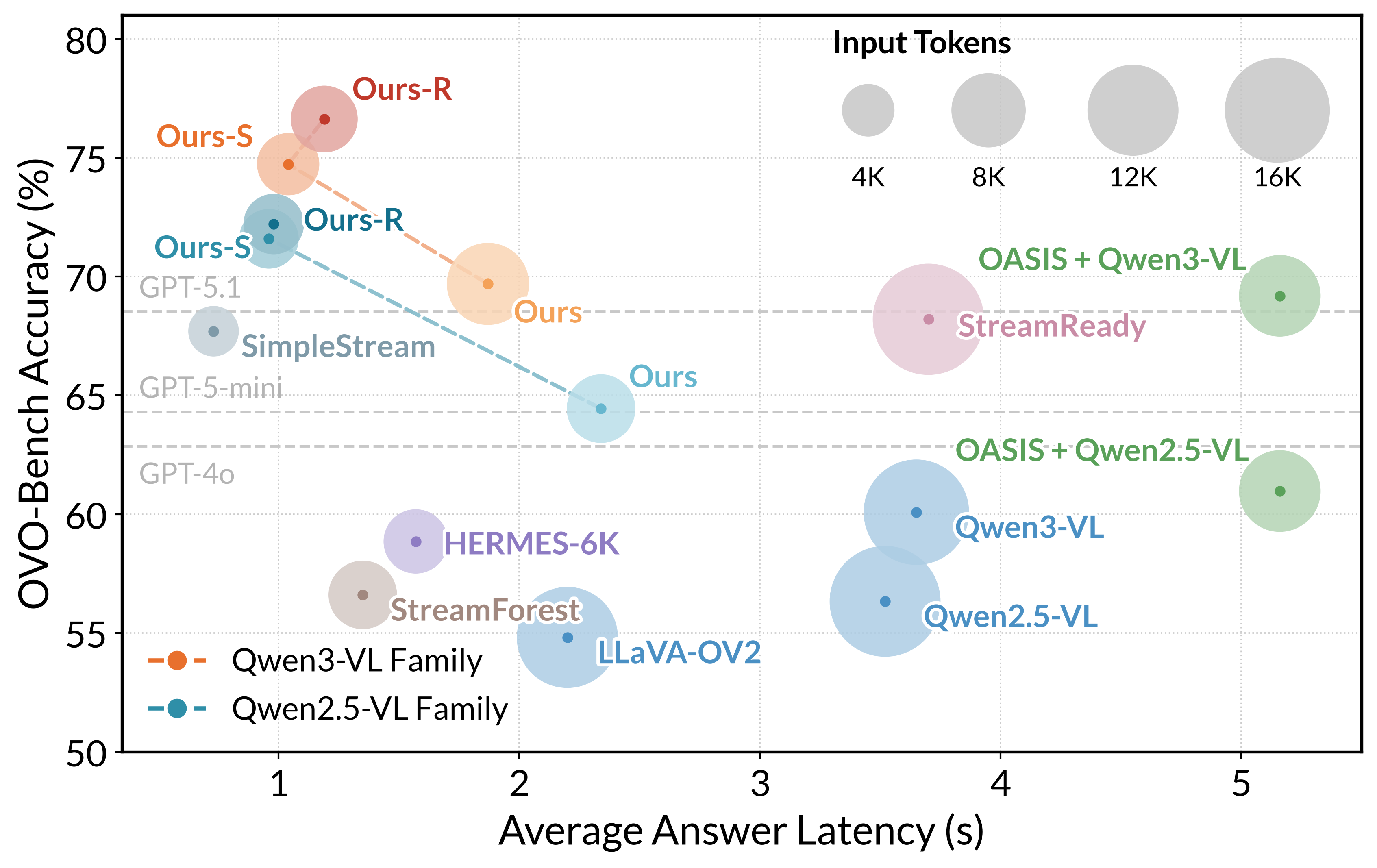}
    \vspace{-18pt}
    \caption{\looseness=-1 \textbf{Accuracy--latency--token trade-off on OVO-Bench.} Marker area is proportional to the average number of input tokens per question. Colors indicate backbone families (\textcolor{q3orange}{\textbf{orange}}: Qwen3-VL; \textcolor{q25blue}{\textbf{blue}}: Qwen2.5-VL), 
    where \emph{Ours}, \emph{Ours-S}, and \emph{Ours-R} denote the zero-shot, supervised, and RL-trained variants, respectively. StreamScout variants achieve stronger accuracy while requiring substantially fewer tokens and lower latency than the baselines.}
    \label{fig:bubble}
    \vspace{-10pt}
\end{figure}

\looseness=-1
Most prior work has devoted most of its effort to the first question, what to retain: a rich body of streaming systems compresses the incoming frames into carefully designed memories, sparse tokens, hierarchical summaries, or retrievable caches~\cite{song2024moviechat,zhang2025flash,di2025streaming,xiong2025streaming,shen2026simple,azad2026streamready,zeng2026streamforest}. Yet however the memory is organized, these systems read it the same way for every question: a fixed amount of context, retrieved by a fixed rule, at a fixed cost. Such uniform access is at odds with the fact that questions are wildly unequal in their demands. For instance, ``\textit{what is the person holding right now?}'' may require only the latest few frames, whereas ``\textit{what price was on the tag shown earlier?}'' requires locating a fine-grained detail buried in the distant past. We therefore argue that \emph{query-adaptive evidence acquisition}, \ie, deciding how deeply to inspect visual evidence for each question, is as important as deciding what the memory stores.

In this work, we introduce \textbf{StreamScout}, a streaming video question-answering framework that makes query-time evidence acquisition an explicit decision of the answering model. As the stream unfolds, StreamScout maintains only a lightweight textual timeline in context. When a question arrives, the model progressively augments this timeline with up to three increasingly informative visual views. At each view, it either answers using the available evidence or explicitly escalates to the next view. The cascade reflects three common locations of evidence in streaming video: a \emph{glance} at recent frames captures the immediate past targeted by many questions; a uniform \emph{look-back} provides coarse coverage of the earlier stream; and query-salient \emph{retrieval} localizes the specific moments most relevant to the question. This design allocates visual evidence according to query difficulty: easy questions terminate after inexpensive inspection, while only questions requiring additional evidence incur the cost of deeper access. Even without post-training, this zero-shot cascade already outperforms prior streaming systems while reducing inference cost, as illustrated in Figure~\ref{fig:bubble}.

The zero-shot model, however, is not always well calibrated: it may stop before sufficient evidence has been acquired or escalate even when the current evidence is already adequate. We sharpen this stop-or-escalate decision using supervision distilled from the model's own behavior. Specifically, we evaluate every stage of the cascade on an auxiliary training set and identify the earliest stage at which the model produces the correct answer. These earliest-successful stages provide direct supervision for the stopping policy and characterize the model's empirical competence boundary, without requiring manual stage annotations or a stronger teacher model. Training a lightweight LoRA on the same interface used during inference yields \textbf{StreamScout-S}. We further optimize the stopping policy through reinforcement learning. Initialized from StreamScout-S, GRPO explores multiple stopping trajectories for each question and jointly rewards answer correctness and lower evidence-acquisition cost. This allows the model to discover effective stopping behaviors beyond those captured by the distilled stage labels, yielding \textbf{StreamScout-R}.

We comprehensively evaluate StreamScout on three standard streaming video understanding benchmarks, OVO-Bench~\cite{niu2025ovo}, StreamingBench~\cite{lin2026streamingbench}, and StreamBench~\cite{xiong2025streaming}, across three widely used backbones: Qwen3-VL~\cite{bai2025qwen3}, Qwen2.5-VL~\cite{bai2025qwen2}, and LLaVA-OneVision-2~\cite{lmms2026llava}. StreamScout and its trained variants consistently surpass prior streaming systems by clear margins, with StreamScout-S sharpening the stop decision and StreamScout-R exploring beyond it, and further exceed proprietary models such as GPT-5.1~\cite{singh2025openai} on OVO-Bench and StreamingBench. StreamScout is also markedly efficient, with StreamScout-S and -R answering with roughly 40\% of the tokens and a quarter of the latency of uniform sampling (Figure~\ref{fig:bubble}).

Our contributions are summarized as follows:
\begin{itemize}[itemsep=2pt, topsep=2pt, leftmargin=*]
    \item We identify query-adaptive evidence acquisition as an overlooked axis of streaming video understanding and propose StreamScout, which dynamically determines how deeply to inspect visual evidence for each question through a cascade of progressively richer views.
    \item \looseness=-1 We further enhance the stop-or-escalate decision with a self-distilled recipe that turns the model's own competence boundary into supervision (StreamScout-S), followed by reinforcement learning that directly optimizes the correctness--cost trade-off through exploration (StreamScout-R).
    \item Extensive experiments on three streaming video benchmarks and three MLLM backbones demonstrate that StreamScout consistently improves accuracy by up to 20.26 percentage points while reducing token consumption by 59\% and answer latency by 72\% on average. 
\end{itemize}

%% file: secs/2_related.tex
\section{Related Work}
\label{sec:related}

\looseness=-1
\textbf{Long-Form Video Understanding.}
Long videos exceed the context of current multi-modal models by orders of magnitude, and a large body of work shrinks them to fit~\cite{tang2025video}. Token- and memory-compression methods condense frames into compact latent representations: MovieChat~\cite{song2024moviechat} pioneered merging dense visual tokens into a sparse long-term memory, Video-XL~\cite{shu2025video} distills hour-long inputs into visual summary tokens, LongVU~\cite{shen2025longvu} adaptively discards spatiotemporally redundant content, and training-free pruning~\cite{tao2025dycoke,zhang2026vscan,shen2026fastvid} cuts visual context at inference time. Another line selects rather than compresses: keyframe sampling methods~\cite{liu2025bolt,tang2025adaptive,ye2025re,zhang2025q} extract a query-relevant subset of frames under a fixed budget. More recently, query-aware methods have grown agentic: VideoAgent~\cite{wang2024videoagent} runs a loop in which an LLM repeatedly judges whether the frames seen so far suffice and requests more, and VideoSeek~\cite{lin2026videoseek} searches for query-relevant moments through multi-round exploration. However, these methods operate strictly offline: the full video can be revisited arbitrarily often, each question costs several model calls, and the sufficiency judgment is often delegated to an external proprietary LLM.

\begin{figure*}
    \centering
    \includegraphics[width=\linewidth]{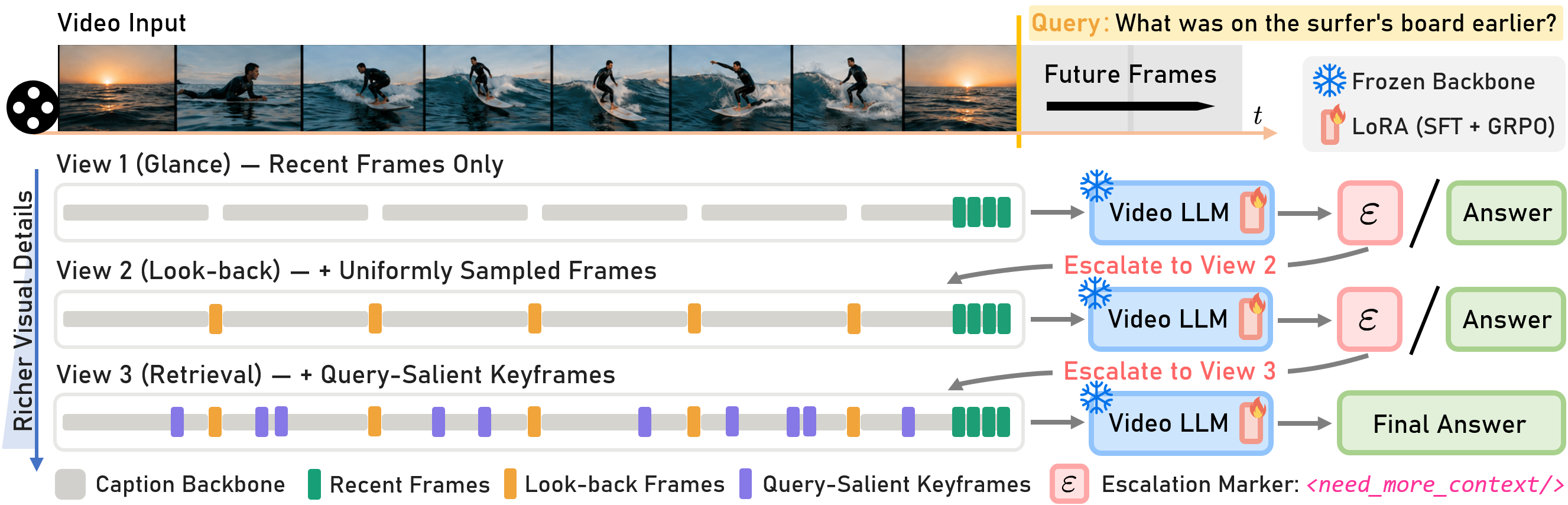}
    \vspace{-17pt}
    \caption{\textbf{Overview of StreamScout.} While the video stream plays, the backbone maintains only a lightweight timestamped caption timeline (gray). A query is answered over the observed prefix through a cascade of up to three progressively richer views: a \textcolor{glancegreen}{\emph{\textbf{glance}}} at the most recent frames, a \textcolor{lookbackorange}{\emph{\textbf{look-back}}} that adds uniformly sampled frames, and a \textcolor{retrievalpurple}{\emph{\textbf{retrieval}}} view that adds query-salient keyframes. At each view the model either commits to an answer or emits the escalation marker \epsbadge{} to request more evidence.}
    \label{fig:pipeline}
    \vspace{-10pt}
\end{figure*}

\smallskip
\looseness=-1
\noindent\textbf{Streaming Video Understanding.}
Streaming understanding adds two constraints: the stream is unbounded and questions arrive at arbitrary timestamps, a setting formalized by recent benchmarks probing real-time perception, backward tracing, and proactive responding~\cite{lin2026streamingbench,niu2025ovo,huang2025online,xiong2025streaming}. Existing systems approach it from two directions. One line trains the model for streaming behavior: VideoLLM-online~\cite{chen2024videollm} interleaves frames with dialogue and learns when to speak, and Dispider~\cite{qian2025dispider} decouples perception from a trained decision of when to respond. The other line designs better memories: ReKV~\cite{di2025streaming} caches and retrieves the KV states of the full stream, HERMES~\cite{zhang2026hermes} balances episodic and semantic memories under a fixed token budget, and FluxMem~\cite{xie2026fluxmem} adaptively updates its memory as the stream evolves; yet however the memory is organized, every question reads it at the same cost. We argue that \emph{how deeply should the system inspect its memory for each individual question} is an equally important yet largely overlooked dimension. 
OASIS~\cite{liang2026oasis} takes a related step by answering on-demand questions through a coarse-to-fine pipeline, but its evidence stages and transition policy are manually specified. In contrast, StreamScout equips the answering model with a learned stop-or-escalate policy over progressively richer evidence views.

%% file: secs/3_method.tex
\vspace{-2pt}
\section{Method}
\label{sec:method}

\looseness=-1
We present StreamScout, a streaming video understanding framework that allocates perception per question through a cascade of progressively richer views, as illustrated in Figure~\ref{fig:pipeline}.

\subsection{Preliminary}
\noindent\textbf{Streaming Video Understanding.}
\looseness=-1
In conventional video question answering, an MLLM answers a query $Q$ over a video of $T$ frames by encoding a frame subset $S$ into visual tokens $\mathbf{x}_v = \mathcal{P}(\mathcal{E}(S))$, interleaving them with the tokenized query $\mathbf{x}_q$, and decoding auto-regressively, $y_j \sim p_{\theta}(y_j \mid \mathbf{x}_v, \mathbf{x}_q, \mathbf{y}_{<j})$; since visual tokens grow linearly with $|S|$ and attention cost quadratically with context length, the number of frames read dominates per-question compute. In the streaming setting, the video is instead an unbounded sequence of frames $\{f_1, f_2, \cdots\}$ arriving in order, and a query $Q$ issued at time $t$ must be answered conditioned solely on the observed prefix $V_{\leq t} = \{f_1, \cdots, f_t\}$. This breaks both assumptions behind offline frame selection. First, since $t$ grows without bound, no fixed token budget can retain the full prefix in context; the system must maintain a compact causal state $\mathcal{M}_t$, updated online as frames arrive. Second, whatever is written into $\mathcal{M}_t$ is written query-agnostically: at ingestion time, future questions are unknown, so no online compression can guarantee to keep what a later question needs.

\subsection{StreamScout: Look Deeper on Demand}
\label{sec:method-StreamScout}
StreamScout separates query-agnostic stream maintenance from query-conditioned visual inspection. As the stream unfolds, it performs fixed-rate processing to construct timestamped captions and maintain a bounded visual cache for subsequent inspection. When a query arrives, the backbone performs fine-grained visual inspection only on a small, adaptively selected set of cached raw frames through a cascade of progressively richer views.

\smallskip
\noindent\textbf{Streaming Memory.}
We segment the incoming stream into consecutive non-overlapping windows of length $\Delta$ and let the backbone itself caption each window from uniformly spaced frames, yielding a timestamped entry
\begin{equation}
c_i = \mathsf{Cap}_{\theta}\big(\{f_j\}_{j \in \mathcal{U}(i)}\big),
\end{equation}
where $\mathcal{U}(i)$ denotes the uniformly spaced frame indices within the $i$-th window. Beyond summarizing what the stream shows, the memory also records what has previously been asked and answered. We initialize the rolling QA summary as $h_0=\varnothing$. After the $j$-th query $Q_j$ is answered at time $t_j$ with committed answer $\hat{a}_j$, the backbone updates the summary as
\begin{equation}
h_{t_j}
=
\mathsf{Fold}_{\theta}
\big(h_{t_j^-},\, Q_j,\, \hat{a}_j\big).
\end{equation}
This rolling summary enables subsequent queries to resolve follow-ups and to reuse previously established conclusions without revisiting the frames from which they were derived. The memory at time $t$ therefore consists of the timestamped caption timeline together with the current QA summary,
\begin{equation}
\mathcal{M}_t \;=\; \Big(\{(\tau_i,\, c_i)\}_{i=1}^{\lfloor t/\Delta \rfloor},\; h_t\Big),
\end{equation}
where $\tau_i$ is the timestamp of the $i$-th window.
Alongside $\mathcal{M}_t$, we maintain a bounded visual cache $\mathcal{C}_t$ of at most $B=64$ previously observed raw frames and their embeddings $\Phi^{\mathsf{vis}}(f_j)$ computed by a lightweight image--text encoder $\Phi$~\cite{radford2021learning}. StreamScout operates in a one-pass manner and never revisits discarded raw frames. When a new frame arrives after the cache is full, we evict the frame in the most temporally dense region, thereby preferentially removing temporally redundant frames while preserving broad coverage of the observed stream.

\smallskip
\noindent\textbf{A Cascade of Views.}
When a query arrives, StreamScout constructs $K{=}3$ progressively richer frame sets $F^{(k)} \subseteq \mathcal{C}_t$:
\begin{align}
F^{(1)} &= \mathsf{Recent}_{n_1}\big(\mathcal{C}_t\big) && \textit{(glance)}, \nonumber\\
F^{(2)} &= F^{(1)} \,\cup\, \mathsf{Back}^{\,\delta}_{n_2}\big(\mathcal{C}_t\big) && \textit{(look-back)}, \\
F^{(3)} &= F^{(2)} \,\cup\, \mathsf{WS}_{n_3}\big(\mathcal{C}_t;\, s\big) && \textit{(retrieval)}, \nonumber
\end{align}
where $\mathsf{Recent}_{n_1}(\cdot)$ selects the $n_1$ most recent cached frames, $\mathsf{Back}^{\,\delta}_{n_2}(\cdot)$ selects up to $n_2$ cached frames approximately at temporal stride $\delta$ counting back from the present, and $\mathsf{WS}_{n_3}(\cdot\,; s)$ selects $n_3$ query-salient frames from the cache under the relevance scores $s_i = \cos\big(\Phi^{\mathsf{vis}}(f_i),\, \Phi^{\mathsf{txt}}(Q)\big),$
computed against the cached frame embeddings. Rather than taking the $n_3$ globally highest-scoring frames, which tend to cluster within a single temporal neighborhood, $\mathsf{WS}$ applies a watershed transform~\cite{haralick1987image} to the relevance curve over the cached frames, partitioning them into temporally distinct segments, and returns the peak frame of the $n_3$ most salient segments. Retrieval thus lands on separate moments rather than repeated shots of the same one. \looseness=-1

Intuitively, the \emph{glance} view covers the immediate past that streaming questions most often target; the \emph{look-back} view extends coarse visual grounding beyond the glance over the cached history; and the \emph{retrieval} view localizes the cached moments most relevant to the question. Since $F^{(1)} \subseteq F^{(2)} \subseteq F^{(3)}$, escalation strictly adds evidence.

Rather than concatenating the selected frames as a separate block, each frame is inserted into the timeline at its timestamped position, producing a single temporally ordered multimodal context $W^{(k)} = \mathsf{Weave}\big(\mathcal{M}_t,\, F^{(k)}\big)$, so that pixel evidence always appears within its textual context and the model can localize what it sees against what it has read. Since every view carries the same textual timeline available at query time, even View~1 retains coarse context over the observed history; the cascade changes only the amount of the cached history that is grounded in raw visual evidence.

\smallskip
\noindent\textbf{Cascaded Inference.}
Given the context $W^{(k)}$, the model first produces one of two decision marker sequences:
$z^{(k)} \in \{\varepsilon, \alpha\}$,
where $\varepsilon=\texttt{<need\_more\_context/>}$ requests the construction of the next view, and $\alpha=\texttt{<answer>}$ commits to answering from the current evidence. If the model emits $\alpha$, it subsequently decodes the answer content $\hat{a}$ autoregressively and terminates with \texttt{</answer>}. If it emits $\varepsilon$, no answer is generated and the system proceeds to view $k{+}1$. At the final view $K$, escalation is disabled and the model must emit $\alpha$.

\subsection{StreamScout-S: Learning the Stop Decision}
\label{sec:method-sft}
\looseness=-1
The cascade above is training-free, yet both its accuracy and its efficiency rest on a single judgment: \emph{is the current evidence sufficient?} A zero-shot model misjudges in both directions, committing prematurely on questions it cannot yet answer and escalating on questions it already could. We show that this judgment can be supervised without human annotation, using the model's own successes and failures as the label source.

\smallskip
\noindent\textbf{Probing the Competence Boundary.}
We take an auxiliary QA set $\mathcal{D}_{\mathrm{aux}}$ with ground-truth answers, disjoint from all evaluation benchmarks, and replay each question $Q \in \mathcal{D}_{\mathrm{aux}}$ through all $K$ views with commitment forced, recording the per-view outcome $o_k(Q) \in \{0, 1\}$, \ie, whether the committed answer matches the ground truth $a^{\star}(Q)$. The shallowest sufficient view,
\begin{equation}
k^{\star}(Q) \;=\; \min \,\{\, k : o_k(Q) = 1 \,\},
\end{equation}
with $k^{\star}(Q) = \infty$ when no view succeeds, is the optimal stopping level for this question under this backbone: stopping earlier yields a wrong answer, and stopping later wastes compute. Questions with $k^{\star} = \infty$ admit no valid stop label and are discarded; this correct-only filtering also prevents the supervision from rewarding confident commitment on questions beyond the model's competence.

\smallskip \looseness=-1
\noindent\textbf{From Probes to Decision Labels.}
The probe outcomes are converted into supervision for the \emph{decision token} only. Let $\alpha=\texttt{<answer>}$ denote the commitment action and $\varepsilon=\texttt{<need\_more\_context/>}$ denote escalation. For each retained question, every view before $k^{\star}$ is labeled with $\varepsilon$, while the shallowest successful view is labeled with $\alpha$:
\begin{equation}
\mathcal{D} \!=\!\bigcup_{Q \,:\, k^{\star}(Q) \,<\, \infty}\!\Big( \big\{ (W^{(k)}, \varepsilon) \big\}_{k < k^{\star}} \cup \big\{ (W^{(k^{\star})},\, \alpha) \big\} \Big).
\end{equation}
Importantly, the ground-truth answer $a^{\star}(Q)$ is used only during probing to determine whether each view succeeds and hence to derive $k^{\star}(Q)$. It is never used as an answer-generation target during supervised training. In particular, all answer-content tokens following \texttt{<answer>} are masked from the loss. The adapter therefore learns whether to stop or escalate, rather than learning to reproduce answers from the auxiliary QA set.

\smallskip
\noindent\textbf{Training Objective.}
We fine-tune a LoRA~\cite{hu2022lora} on the language-tower projections only, keeping the vision tower and projector frozen. Let $\tilde{\theta}$ denote the backbone parameters augmented with the adapter; the objective is the token-level negative log-likelihood over the target token
\begin{equation}
\mathcal{L}(\tilde{\theta}) \;=\; -\!\!\sum_{(W,\, z) \in \mathcal{D}} \log p_{\tilde{\theta}}\big(z \mid W\big).
\end{equation}

\input{tables/ovo}

\input{tables/streambench}
\subsection{StreamScout-R: Reinforcing the Stop Decision}
\label{sec:method-rl}
\looseness=-1
StreamScout-S imitates the earliest empirically successful view, but this per-view supervision never evaluates a trajectory as a whole: whether an escalation is worth taking depends jointly on the answer it eventually enables and the cost it incurs. We therefore optimize the stop-or-escalate policy directly against this trajectory-level trade-off, using reinforcement learning initialized from StreamScout-S.

\smallskip
\noindent\textbf{Decision Rollouts.}
For a query $(Q, t)$ we roll out the cascade under the current policy: at each view the model samples either the escalation marker $\varepsilon$ or a committed answer, and the trajectory terminates at the committed view $\hat{k}$. Because the woven contexts $W^{(1)},\dots,W^{(K)}$ are deterministic given $\mathcal{M}_t$, an entire trajectory is determined by the model's view-level choices, so rollouts reuse the cached views and never re-perceive frames. Each trajectory $o$ receives a scalar reward that couples correctness with the perception it spent,
\begin{equation}
R(o) \;=\; \underbrace{\big(\mathds{1}[\hat{a}=a^{\star}] - \eta\,\mathds{1}[\hat{a}\neq a^{\star}]\big)}_{\text{correctness}} \;-\; \lambda\,\underbrace{\frac{\hat{k}-1}{K-1}}_{\text{cost}},
\end{equation}
where $\hat{a}$ is the committed answer, $\eta$ penalizes a wrong commitment so that guessing carries negative expected value, and the normalized cost term charges each escalation, scaled by $\lambda$. The coefficient $\lambda$ is an explicit knob on the accuracy--compute trade-off: larger $\lambda$ buys cheaper trajectories, smaller $\lambda$ buys deeper ones. Because the reward depends only on a verifiable answer match, no reward model is needed.

\smallskip
\noindent\textbf{Group-Relative Policy Optimization.}
We optimize with GRPO~\cite{shao2024deepseekmath}. For each query we draw a group of $G$ trajectories $\{o_i\}_{i=1}^{G}$ from the behavior policy and form the advantage of each by centering its reward on the group mean,
\begin{equation}
A_i \;=\; R(o_i) \;-\; \frac{1}{G}\sum_{j=1}^{G} R(o_j).
\end{equation}
Following Dr.\,GRPO~\cite{liu2025understanding}, we omit the per-group standard-deviation normalization of the original estimator: on this task a large fraction of groups are near-degenerate, all trajectories right or all wrong, and dividing by a vanishing spread rescales the $\lambda$-sized cost differences within such groups up to the full magnitude of the correctness signal, which drives the policy to collapse onto always answering at View~1. We further skip any group whose rewards are identical, for which every advantage is zero. The retained groups update the adapter through the clipped, token-level surrogate
\begin{equation}
\begin{aligned}
\mathcal{J}(\theta) = {}& \mathbb{E}\Big[\, \textstyle\sum_{i,u} \min\big(\rho_{i,u}A_i,\; \mathrm{clip}(\rho_{i,u}, 1\,{\pm}\,\epsilon)\,A_i\big)\Big] \\[2pt]
& - \beta\,\mathbb{D}_{\mathrm{KL}}\!\big(\pi_\theta \,\|\, \pi_{\mathrm{ref}}\big),
\end{aligned}
\end{equation}
where $\rho_{i,u}=\pi_\theta(o_{i,u})/\pi_{\theta_{\mathrm{old}}}(o_{i,u})$ is the per-token importance ratio and the KL term is estimated with the unbiased $k_3$ estimator. Crucially, the reference $\pi_{\mathrm{ref}}$ is the StreamScout-S policy itself. In practice we realize $\pi_{\mathrm{ref}}$ for free by merging the StreamScout-S adapter into the backbone and training a fresh adapter on top, so that disabling the new adapter recovers the reference in place.

%% file: tables/ovo.tex
\begin{table*}[t]
    \centering
    \setlength{\tabcolsep}{8pt}
    \caption{\looseness=-1 \textbf{Performance comparison on OVO-Bench}. 
    Colors on model names indicate categories: \legendbox{cprop}{proprietary models}, \legendbox{coffline}{open-source offline MLLMs}, and \legendbox{conline}{open-source online MLLMs}. \textbf{Avg} and \textbf{Overall} are question-weighted accuracies over each panel's questions and all 1{,}468 questions, respectively. Within each backbone block, the best score in each column is in \textbf{bold} and the second best is \underline{underlined}. $^\dagger$ denotes on-demand methods whose per-question frame count is adaptive.}
    \vspace{-5pt}
    \label{tab:ovo-bench}
    \resizebox{\textwidth}{!}{%
    \begin{tabular}{lccccccccccccc}
    \toprule
    \multirow{2}{*}{\textbf{Model}} & \multirow{2}{*}{\textbf{Frames}} & \multicolumn{7}{c}{\textbf{Real-Time Visual Perception}} & \multicolumn{4}{c}{\textbf{Backward Tracing}} & \multirow{2}{*}{\textbf{Overall}} \\
    \cmidrule(lr){3-9}\cmidrule(lr){10-13}
    & & OCR & ACR & ATR & STU & FPD & OJR & \textbf{Avg} & EPM & ASI & HLD & \textbf{Avg} & \\
    \midrule
    \cellcolor{cprop}Gemini 1.5 Pro & 1 fps & 85.91 & 66.97 & 79.31 & 58.43 & 63.37 & 61.96 & 68.70 & 58.59 & 76.35 & 52.64 & 61.00 & 65.39 \\
    \cellcolor{cprop}GPT-4o & 64 & 69.80 & 64.22 & 71.55 & 51.12 & 70.30 & 59.78 & 63.20 & 57.91 & 75.68 & 48.66 & 59.35 & 61.55 \\
    \cellcolor{cprop}GPT-5-mini & 64 & 81.21 & 67.89 & 81.90 & 66.29 & 73.27 & 66.30 & 72.16 & 66.67 & 75.68 & 16.13 & 53.89 & 64.31 \\
    \cellcolor{cprop}GPT-5.1 & 64 & 83.22 & 78.90 & 77.59 & 67.42 & 79.21 & 76.63 & 76.58 & 61.95 & 73.65 & 38.71 & 57.84 & 68.53 \\
    \cellcolor{coffline}LLaVA-Video-7B & 64 & 69.13 & 58.72 & 68.83 & 49.44 & 74.26 & 59.78 & 62.11 & 56.23 & 57.43 & 7.53 & 42.16 & 53.53 \\
    \cellcolor{coffline}LLaVA-OneVision-7B & 64 & 66.44 & 57.80 & 73.28 & 53.37 & 71.29 & 61.96 & 63.08 & 54.21 & 55.41 & 21.51 & 44.85 & 55.25 \\
    \cellcolor{coffline}Qwen2-VL-7B & 64 & 60.40 & 50.46 & 56.03 & 47.19 & 66.34 & 55.43 & 55.31 & 47.81 & 35.48 & 56.08 & 47.36 & 51.89 \\
    \cellcolor{coffline}Qwen2-VL-72B & 64 & 65.77 & 60.55 & 69.83 & 51.69 & 69.31 & 54.35 & 60.58 & 52.53 & 60.81 & 57.53 & 55.95 & 58.59 \\
    \cellcolor{conline}Flash-VStream-7B & 1 fps & 24.16 & 29.36 & 28.45 & 33.71 & 25.74 & 28.80 & 28.67 & 39.06 & 37.16 & 5.91 & 28.84 & 28.75 \\
    \cellcolor{conline}VideoLLM-online-8B & 2 fps & 8.05 & 23.85 & 12.07 & 14.04 & 45.54 & 21.20 & 19.35 & 22.22 & 18.80 & 12.18 & 18.46 & 18.97 \\
    \cellcolor{conline}StreamChat & 15 fps & 51.68 & 47.71 & 58.62 & 41.01 & 61.06 & 47.83 & 50.14 & 50.17 & 54.05 & 37.63 & 47.38 & 48.96 \\
    \cellcolor{conline}Dispider & 1 fps & 57.72 & 49.54 & 62.07 & 44.94 & 61.39 & 51.63 & 53.64 & 48.48 & 55.41 & 4.30 & 37.08 & 46.53 \\
    \midrule
    Qwen2.5-VL-7B & 0.5 fps & 76.51 & 57.80 & 68.10 & 46.63 & 66.34 & 56.52 & 60.93 & 49.83 & 58.11 & 44.62 & 50.24 & 56.33 \\
    + HERMES (6K) & 0.5 fps & 85.91 & 60.55 & 74.14 & 52.81 & 70.30 & 66.85 & 67.86 & 49.49 & 58.78 & 33.33 & 46.91 & 58.85 \\
    + HERMES (4K) & 0.5 fps & 85.23 & 64.22 & 71.55 & 53.37 & 74.26 & 65.22 & 68.10 & 48.48 & 62.16 & 37.63 & 48.49 & 59.67 \\
    + FluxMem & 1 fps & 81.21 & 59.63 & 70.69 & 53.37 & \textbf{75.25} & 63.04 & 66.31 & 48.48 & \textbf{64.19} & 29.03 & 46.43 & 57.76 \\
    + OASIS & 48\,/\,80$^\dagger$ & 85.23 & 72.48 & 66.38 & 52.25 & 67.33 & 64.67 & 67.26 & 51.85 & 58.78 & 48.92 & 52.61 & 60.97 \\
    \rowcolor{gray!10}
    \textbf{+ StreamScout} & 4\,/\,12\,/\,24$^\dagger$ & 91.95 & \underline{77.06} & 74.14 & \underline{58.99} & 74.26 & \underline{77.72} & 75.27 & 53.54 & 60.14 & 36.56 & 50.08 & 64.44 \\
    \rowcolor{gray!10}
    \textbf{+ StreamScout-S} & 4\,/\,12\,/\,24$^\dagger$ & \textbf{93.29} & 74.31 & \textbf{77.59} & \textbf{59.55} & 74.26 & 76.63 & \underline{75.51} & \underline{53.87} & \textbf{64.19} & \underline{88.17} & \underline{66.40} & \underline{71.59} \\
    \rowcolor{gray!10}
    \textbf{+ StreamScout-R} & 4\,/\,12\,/\,24$^\dagger$ & \underline{92.62} & \textbf{77.98} & \textbf{77.59} & 57.30 & \textbf{75.25} & \textbf{78.26} & \textbf{75.87} & \textbf{55.22} & \textbf{64.19} & \textbf{89.25} & \textbf{67.35} & \textbf{72.21} \\
    \midrule
    Qwen3-VL-8B & 0.5 fps & 83.89 & 58.72 & 70.69 & 56.18 & 69.31 & 64.13 & 66.79 & 51.18 & 60.81 & 43.55 & 51.19 & 60.08 \\
    + OASIS & 48\,/\,80$^\dagger$ & \underline{92.62} & 81.65 & 78.45 & \underline{65.17} & 68.00 & 78.26 & 77.26 & 58.25 & 67.57 & 51.61 & 58.48 & 69.19 \\
    \rowcolor{gray!10}
    \textbf{+ StreamScout} & 4\,/\,12\,/\,24$^\dagger$ & 91.28 & \underline{82.57} & \textbf{82.76} & 64.04 & \textbf{75.25} & 75.00 & 77.66 & 55.22 & 61.49 & 63.44 & 59.11 & 69.69 \\
    \rowcolor{gray!10}
    \textbf{+ StreamScout-S} & 4\,/\,12\,/\,24$^\dagger$ & \underline{92.62} & \underline{82.57} & \underline{81.03} & \underline{65.17} & 73.27 & \textbf{79.35} & \underline{78.62} & \textbf{60.94} & \underline{68.24} & \underline{84.41} & \underline{69.57} & \underline{74.73} \\
    \rowcolor{gray!10}
    \textbf{+ StreamScout-R} & 4\,/\,12\,/\,24$^\dagger$ & \textbf{93.96} & \textbf{84.40} & 80.17 & \textbf{67.42} & \underline{74.26} & \underline{78.80} & \textbf{79.45} & \underline{58.59} & \textbf{69.59} & \textbf{98.39} & \textbf{72.90} & \textbf{76.63} \\
    \midrule
    LLaVA-OneVision-2-8B & 0.5 fps & 62.42 & 62.39 & 68.97 & 61.80 & 62.00 & 54.89 & 61.49 & 53.20 & 62.84 & 20.97 & 45.96 & 54.81 \\
    + OASIS & 48\,/\,80$^\dagger$ & 80.54 & 59.63 & 61.21 & 52.25 & 71.29 & 59.78 & 63.44 & 46.46 & 56.08 & 39.78 & 46.75 & 56.27 \\
    \rowcolor{gray!10}
    \textbf{+ StreamScout} & 4\,/\,12\,/\,24$^\dagger$ & \underline{88.59} & 77.98 & \textbf{80.17} & 65.73 & 73.27 & 75.54 & 76.46 & 54.55 & 60.81 & 73.66 & 61.65 & 70.10 \\
    \rowcolor{gray!10}
    \textbf{+ StreamScout-S} & 4\,/\,12\,/\,24$^\dagger$ & \underline{88.59} & \underline{79.82} & \underline{79.31} & \textbf{69.66} & \textbf{75.25} & \underline{78.80} & \underline{78.37} & \textbf{62.29} & \underline{70.27} & \underline{84.41} & \underline{70.68} & \underline{75.07} \\
    \rowcolor{gray!10}
    \textbf{+ StreamScout-R} & 4\,/\,12\,/\,24$^\dagger$ & \textbf{89.93} & \textbf{80.73} & \underline{79.31} & \underline{67.98} & \textbf{75.25} & \textbf{79.35} & \textbf{78.49} & \underline{61.62} & \textbf{72.30} & \textbf{90.81} & \textbf{72.73} & \textbf{76.02} \\
    \bottomrule
    \end{tabular}%
    }
    \vspace{-10pt}
\end{table*}

%% file: tables/streambench.tex
\begin{table*}[t]
    \centering
    \setlength{\tabcolsep}{10pt}
    \caption{\textbf{Performance comparisons on StreamingBench and StreamBench}. 
    Colors on model names indicate categories: \legendbox{cprop}{proprietary models}, \legendbox{coffline}{open-source offline MLLMs}, and \legendbox{conline}{open-source online MLLMs}. 
    Within each backbone block, the best score in each column is in \textbf{bold} and the second best is \underline{underlined}. 
    $^\dagger$ denotes adaptive per-question frame count.}
    \vspace{-5pt}
    \label{tab:stream_results}
    \resizebox{\textwidth}{!}{%
    \begin{tabular}{l c ccccc ccccccc}
    \toprule
    \multirow{2}{*}{\textbf{Model}} & \multirow{2}{*}{\textbf{Frames}} & \multicolumn{5}{c}{\textbf{StreamingBench}} & \multicolumn{7}{c}{\textbf{StreamBench}} \\
    \cmidrule(lr){3-7}\cmidrule(lr){8-14}
    & & Real-Time & ACU & MCU & SQA & \textbf{Avg} & OS & LM & SM & CI & KG & SF & \textbf{Avg} \\
    \midrule
    \cellcolor{cprop}Gemini 1.5 Pro      & 1 fps     & 75.7 & 51.4 & 40.7 & 54.8 & 69.5 & --   & --   & --   & --   & --   & --   & --   \\
    \cellcolor{cprop}GPT-4o              & 64 & 73.3 & 41.2 & 38.4 & 32.8 & 65.0 & 60.5 & 61.2 & 64.4 & 72.3 & 93.9 & 74.7 & 71.0 \\
    \cellcolor{cprop}GPT-5-mini          & 64 & 82.6 & 46.0 & 52.8 & 54.4 & 75.3 & 68.6 & 75.6 & 73.2 & 79.8 & 64.4 & 79.8 & 73.6 \\
    \cellcolor{cprop}GPT-5.1             & 64 & 83.5 & 49.6 & 51.2 & 52.4 & 76.0 & 67.2 & 67.0 & 72.0 & 79.5 & 66.8 & 82.1 & 72.4 \\
    \cellcolor{coffline}LongVA              & 8         & --   & --   & --   & --   & --   & 41.1 & 47.4 & 57.6 & 59.8 & 80.7 & 66.1 & 52.4 \\
    \cellcolor{coffline}InternVL2-8B        & 16  & 63.7 & 32.0 & 31.2 & 32.3 & 56.3 & 38.5 & 46.6 & 50.9 & 67.6 & 81.0 & 62.2 & 57.6 \\
    \cellcolor{coffline}Qwen2-VL-7B         & 0.2--1 fps & 69.0 & 31.2 & 26.0 & 39.6 & 60.5 & -- & -- & -- & -- & -- & -- & -- \\
    \cellcolor{coffline}Video-LLaMA2-7B     & 32        & 49.5 & 24.8 & 26.8 & 18.7 & 43.5 & --   & --   & --   & --   & --   & --   & --   \\
    \cellcolor{conline}MovieChat           & 32        & --   & --   & --   & --   & --   & 18.6 & 20.4 & 26.5 & 42.3 & 67.2 & 35.8 & 35.3 \\
    \cellcolor{conline}VideoLLM-online     & 5 fps     & --   & --   & --   & --   & --   & 41.4 & 48.8 & 52.9 & 62.7 & 69.2 & 64.1 & 56.4 \\
    \cellcolor{conline}Flash-VStream       & 1 fps     & --   & --   & --   & --   & --   & 37.1 & 44.5 & 48.6 & 58.1 & 66.4 & 59.2 & 52.1 \\
    \cellcolor{conline}StreamChat & 15 fps  & 60.1 & 28.8 & 22.4 & 32.0 & 52.6 & 40.7 & 43.6 & 47.1 & 63.0 & 89.1 & 73.8 & 59.5 \\
    \midrule
    Qwen2.5-VL-7B & 0.5 fps & 66.0 & 33.2 & 30.1 & 34.0 & 58.3 & 36.1 & 41.3 & 41.0 & 50.5 & 76.5 & 62.3 & 51.1 \\
    + OASIS & 48\,/\,80$^\dagger$ & 70.6 & 38.8 & 36.0 & 46.0 & 63.6 & 35.9 & 39.4 & 44.1 & 54.5 & 76.9 & 64.9 & 52.4 \\
    \rowcolor{gray!10}
    \textbf{+ StreamScout} & 4\,/\,12\,/\,24$^\dagger$ & 77.7 & \textbf{58.0} & 39.2 & 53.2 & 71.3 & 34.8 & 32.0 & 43.5 & 41.4 & 83.6 & 63.8 & 49.7 \\
    \rowcolor{gray!10}
    \textbf{+ StreamScout-S} & 4\,/\,12\,/\,24$^\dagger$ & \underline{78.9} & \underline{54.8} & \underline{43.2} & \textbf{58.4} & \textbf{72.7} & \underline{41.1} & \textbf{48.5} & \underline{53.8} & \underline{60.3} & \textbf{89.3} & \textbf{74.8} & \underline{61.1} \\
    \rowcolor{gray!10}
    \textbf{+ StreamScout-R} & 4\,/\,12\,/\,24$^\dagger$ & \textbf{79.3} & 54.4 & \textbf{44.0} & \underline{54.0} & \textbf{72.7} & \textbf{41.5} & \textbf{48.5} & \textbf{55.5} & \textbf{64.3} & \underline{86.2} & \underline{73.8} & \textbf{61.5} \\
    \midrule
    Qwen3-VL-8B & 0.5 fps & 72.8 & 35.6 & 35.7 & 43.9 & 64.9 & 33.8 & 50.5 & 47.8 & 69.4 & 73.5 & 62.6 & 56.0 \\
    + OASIS & 48\,/\,80$^\dagger$ & 78.2 & 42.7 & 49.6 & 48.4 & 71.0 & 47.0 & 53.6 & 54.3 & \textbf{71.7} & \textbf{86.1} & 67.6 & 62.1 \\
    \rowcolor{gray!10}
    \textbf{+ StreamScout} & 4\,/\,12\,/\,24$^\dagger$ & 80.3 & 48.0 & 42.0 & 42.0 & 71.9 & \underline{52.5} & 51.8 & \underline{61.4} & \underline{71.4} & \underline{84.6} & 70.8 & \underline{65.3} \\
    \rowcolor{gray!10}
    \textbf{+ StreamScout-S} & 4\,/\,12\,/\,24$^\dagger$ & \textbf{82.8} & \underline{55.2} & \underline{52.0} & \textbf{54.0} & \underline{76.1} & \underline{52.5} & \textbf{58.8} & 57.5 & 69.4 & 77.9 & \underline{71.2} & 64.4 \\
    \rowcolor{gray!10}
    \textbf{+ StreamScout-R} & 4\,/\,12\,/\,24$^\dagger$ & \textbf{82.8} & \textbf{59.6} & \textbf{53.6} & \underline{53.2} & \textbf{76.5} & \textbf{55.5} & \underline{55.5} & \textbf{61.7} & \underline{71.4} & 76.2 & \textbf{77.4} & \textbf{66.1} \\
    \bottomrule
    \end{tabular}%
    }
    \vspace{-10pt}
\end{table*}

%% file: secs/4_experiments.tex
\section{Experiments}
\label{sec:exp}
In this section, we validate the effectiveness of StreamScout across three different backbones and various streaming video question answering benchmarks.

\subsection{Experimental Settings}
\looseness=-1
\textbf{Backbones and Benchmarks.}
To demonstrate the effectiveness of our approach, we instantiate StreamScout on three backbones spanning different architectures and vision towers: Qwen3-VL-8B~\cite{bai2025qwen3}, Qwen2.5-VL-7B~\cite{bai2025qwen2}, and LLaVA-OneVision-2-8B~\cite{lmms2026llava}. We evaluate on three streaming video understanding benchmarks:
\begin{itemize}[itemsep=2pt, topsep=2pt, leftmargin=*]
    \item \textbf{OVO-Bench~\cite{niu2025ovo}} comprises 2,814 questions over 644 videos, each issued at an annotated timestamp in the stream; we evaluate on its nine real-time visual perception and backward tracing tasks (1,468 questions);
    \item \looseness=-1\textbf{StreamingBench~\cite{lin2026streamingbench}} poses 4,500 questions over 900 videos, spanning 18 tasks from object and action perception to counting, causal reasoning, \etc;
    \item \looseness=-1\textbf{StreamBench~\cite{xiong2025streaming}} provides 1,838 free-form questions over 307 egocentric, web, and movie videos, spanning question types from short-horizon perception to long-horizon memory, with answers scored by an LLM judge.
\end{itemize}
Additionally, we evaluate whether StreamScout transfers to conventional offline video understanding on NExT-QA~\cite{xiao2021next}, EgoSchema~\cite{mangalam2023egoschema}, and MLVU~\cite{zhou2025mlvu}, where StreamScout simply treats each video as a stream.
Collectively, these benchmarks pose questions whose supporting evidence ranges from the most recent frames to details that appeared earlier, and they span both multiple-choice and open-ended answer formats.\footnote{For the streaming benchmarks, videos are accessed and processed online as streams during evaluation, with only the currently available content consumed by the system; the full videos are not downloaded or persistently stored locally.}

\smallskip
\noindent\textbf{Baselines.}
For each backbone, we report a uniform-sampling baseline, an offline recipe that samples the stream at 0.5\,fps up to 64 frames and answers in a single pass. We further compare against representative streaming systems, including VideoLLM-online~\cite{chen2024videollm}, Flash-VStream~\cite{zhang2025flash}, Dispider~\cite{qian2025dispider}, StreamChat~\cite{xiong2025streaming}, HERMES~\cite{zhang2026hermes}, FluxMem~\cite{xie2026fluxmem}, and OASIS~\cite{liang2026oasis}.

\smallskip
\noindent\textbf{Implementation Details}. As the stream unfolds, the backbone itself captions every segment of $\Delta = 16$ seconds, forming the textual timeline. At query time, View~1 weaves the $n_1 = 4$ most recent frames into the timeline; View~2 adds up to $n_2 = 8$ frames at stride $\delta = 32$s counting back from the present; View~3 adds $n_3 = 12$ query-salient frames retrieved with CLIP ViT-L/14~\cite{radford2021learning}. Escalation is triggered by the model emitting the dedicated request token, and the final view ($K \leq 3$) forces a commitment. For StreamBench, all methods are scored by the same LLaMA-3~\cite{grattafiori2024llama} judge to ensure comparability. 
Further training details are provided in the Appendix.

\subsection{Experimental Results}
\looseness=-1
\noindent\textbf{Results on OVO-Bench.}
Table~\ref{tab:ovo-bench} compares StreamScout with existing streaming systems on OVO-Bench. Without any training, StreamScout already establishes state-of-the-art results, reaching 64.44 with Qwen2.5-VL-7B and 69.69 with Qwen3-VL-8B, surpassing all prior streaming approaches on both backbones with at most 24 frames per question, whereas prior systems read their full memory regardless of question difficulty. The cascade also generalizes beyond the Qwen family: with LLaVA-OneVision-2-8B it reaches 70.10 zero-shot and delivers the largest gain of 13.8\% over OASIS among the three backbones. Learning the decision pushes further: StreamScout-S adds 7.15, 5.04, and 4.97 points across the three backbones, with the largest gains on Backward Tracing, where assessing evidence sufficiency is challenging. StreamScout-R further raises the best overall accuracy to 72.21 and 76.63 through exploration, exceeding GPT-5.1 with 8B-scale models. Overall, StreamScout consistently delivers state-of-the-art performance, and its sufficiency judgment is progressively resolved by self-distillation and reinforcement learning.

\smallskip
\noindent\textbf{Results on StreamingBench.}
As shown in Table~\ref{tab:stream_results}, the zero-shot cascade alone lifts the question-weighted average from 58.3 to 71.3 with Qwen2.5-VL-7B and from 64.9 to 71.9 with Qwen3-VL-8B, surpassing all prior streaming systems on both backbones. The gain is most pronounced on ACU, where accuracy rises by up to 24.8 points, indicating that the recency and retrieval views together capture transient anomalies that no fixed sampling rate can anticipate. Training pushes further: StreamScout-S recovers the two subsets that stress the sufficiency judgment most, MCU and SQA, and StreamScout-R attains the best averages of 72.7 and 76.5 on the two backbones, with the Qwen3-VL variant exceeding GPT-5.1 and both exceeding Gemini 1.5 Pro (69.5).

\smallskip
\noindent\textbf{Results on StreamBench.}
Table~\ref{tab:stream_results} further evaluates whether these gains transfer to open-ended video question answering, where responses are free-form and scored by an LLM judge. StreamScout-R achieves the best average on both backbones, reaching 61.5 with Qwen2.5-VL-7B and 66.1 with Qwen3-VL-8B, and surpassing OASIS and all other streaming systems. These results show that StreamScout generalizes beyond multiple-choice benchmarks to free-form answering.

\smallskip
\noindent\textbf{Results on Offline Benchmarks.}
Table~\ref{tab:offline} evaluates whether the streaming recipe transfers to conventional offline video QA, where StreamScout simply treats each video as a stream; none of these benchmarks appear in any training data. The zero-shot cascade trades below the uniform-sampling baseline since the offline questions are not biased toward the recent frames. Training recovers and surpasses it: StreamScout-S and StreamScout-R exceed the base model on three benchmarks, while reading a smaller number of frames per question. On these offline benchmarks, the learned variants more often escalate to the query-salient retrieval view, which can locate sparse evidence anywhere in the video.

\begin{table}[t]
  \centering
  \small
  \setlength{\tabcolsep}{6pt}
  \caption{\textbf{Performance on offline video benchmarks with Qwen3-VL-8B.}  $^\dagger$~denotes adaptive per-question frame count.}
  \vspace{-7pt}
  \label{tab:offline}
  \resizebox{\linewidth}{!}{%
  \begin{tabular}{lcccc}
    \toprule
    \textbf{Method} & \textbf{Frames} & \textbf{NExT-QA} & \textbf{EgoSchema} & \textbf{MLVU} \\
    \midrule
    Qwen3-VL-8B     & 64 & 75.68 & 72.60  & 67.53\\
    + OASIS   & 48 / 80$^\dagger$ & 72.23 & 69.00 & 59.79\\
    \rowcolor{gray!10}+ StreamScout   & 4\,/\,12\,/\,24$^\dagger$ & 72.75 & 68.60 & 63.03\\
    \rowcolor{gray!10}+ StreamScout-S & 4\,/\,12\,/\,24$^\dagger$ & 77.18 & 73.40 & 69.78\\
    \rowcolor{gray!10}+ StreamScout-R & 4\,/\,12\,/\,24$^\dagger$ & \textbf{78.41} & \textbf{74.60} & \textbf{71.74}\\
    \bottomrule
  \end{tabular}
  }
\end{table}

\subsection{Discussions}
\noindent\textbf{Ablation Studies.}
Table~\ref{tab:view-ablation} ablates the frame sources used by zero-shot StreamScout on OVO-Bench while retaining the textual timeline in all configurations. The glance view alone performs remarkably well, confirming that recent frames are sufficient for a large proportion of streaming questions. In contrast, the look-back view alone drops to 44.56, suggesting that uniform temporal coverage is too coarse to support direct answering. The pairwise variants reveal complementary roles among the auxiliary views. Glance provides the strongest foundation for perception, while adding look-back yields a modest further improvement on perception. Retrieval, by contrast, contributes most to backward tracing, recovering distant details that may be abstracted away in the caption-based timeline. Combining all three views captures both benefits and yields the best overall accuracy of 69.69\%.

\begin{table}[t]
    \centering
    \small
    \caption{\textbf{View-recipe ablation on OVO-Bench}. Each column enables a subset of frame sources. Multi-view recipes keep the stop-or-escalate cascade; single-view recipes answer directly.}
    \vspace{-7pt}
    \setlength{\tabcolsep}{6pt}
    \label{tab:view-ablation}
  \resizebox{\linewidth}{!}{%
    \begin{tabular}{@{}lccccccc}
    \toprule
    \textbf{Views} & V1 & V2 & V3 & V1{+}2 & V1{+}3 & V2{+}3 & \cellcolor{gray!10}Full \\
    \midrule
    Perception   & 77.24 & 44.91 & 66.47 & \textbf{77.84} & 76.44 & 67.07 & \cellcolor{gray!10}\underline{77.66} \\
    Backward   & 52.12 & 44.09 & 53.54 & 53.54 & \underline{58.63} & 54.33 & \cellcolor{gray!10}\textbf{59.11} \\
    Overall & 66.39 & 44.56 & 60.88 & 67.35 & \underline{68.75} & 61.56 & \cellcolor{gray!10}\textbf{69.69} \\
    \bottomrule
    \end{tabular}
    }
    \vspace{-5pt}
\end{table}

\begin{figure}[t]
    \centering
    \begin{minipage}[c]{0.035\linewidth}
        \centering
        \rotatebox{90}{\scriptsize (a) Answer Latency (s)}
    \end{minipage}%
    \begin{minipage}[c]{0.96\linewidth}
        \includegraphics[width=\linewidth]{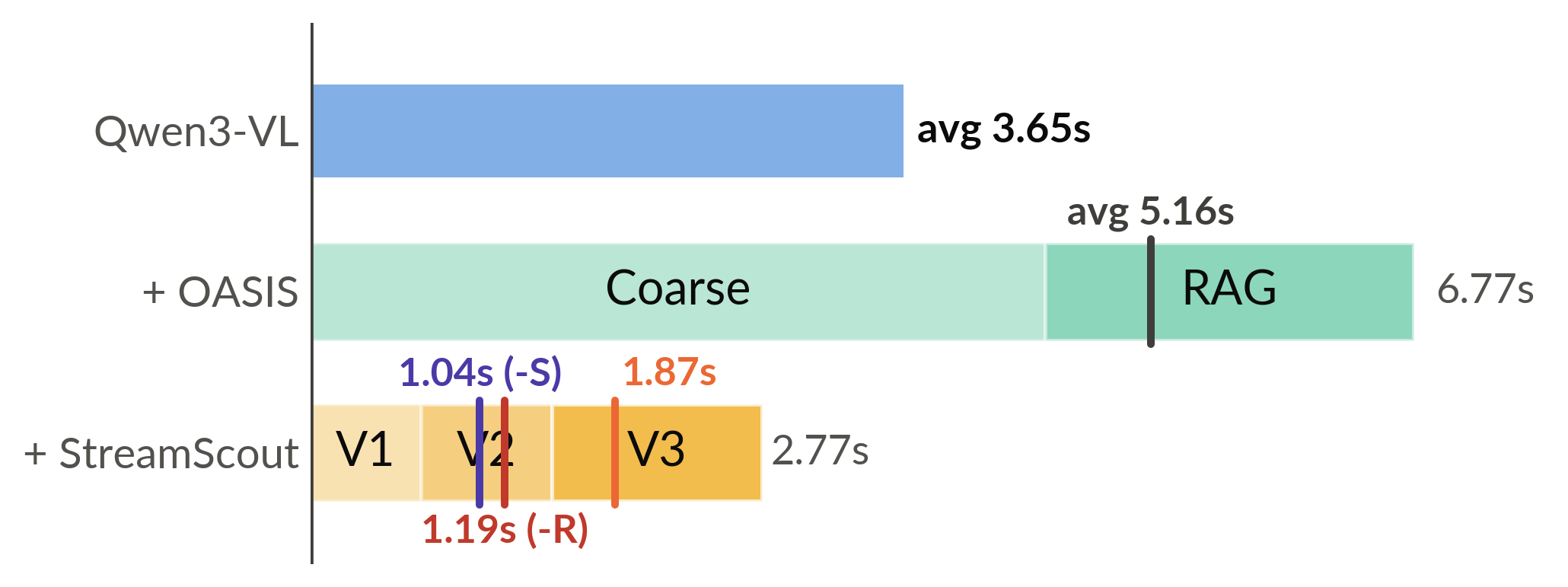}
    \end{minipage}\\[-2pt]
    \begin{minipage}[c]{0.035\linewidth}
        \centering
        \rotatebox{90}{\scriptsize (b) Token Consumption}
    \end{minipage}%
    \begin{minipage}[c]{0.96\linewidth}
        \includegraphics[width=\linewidth]{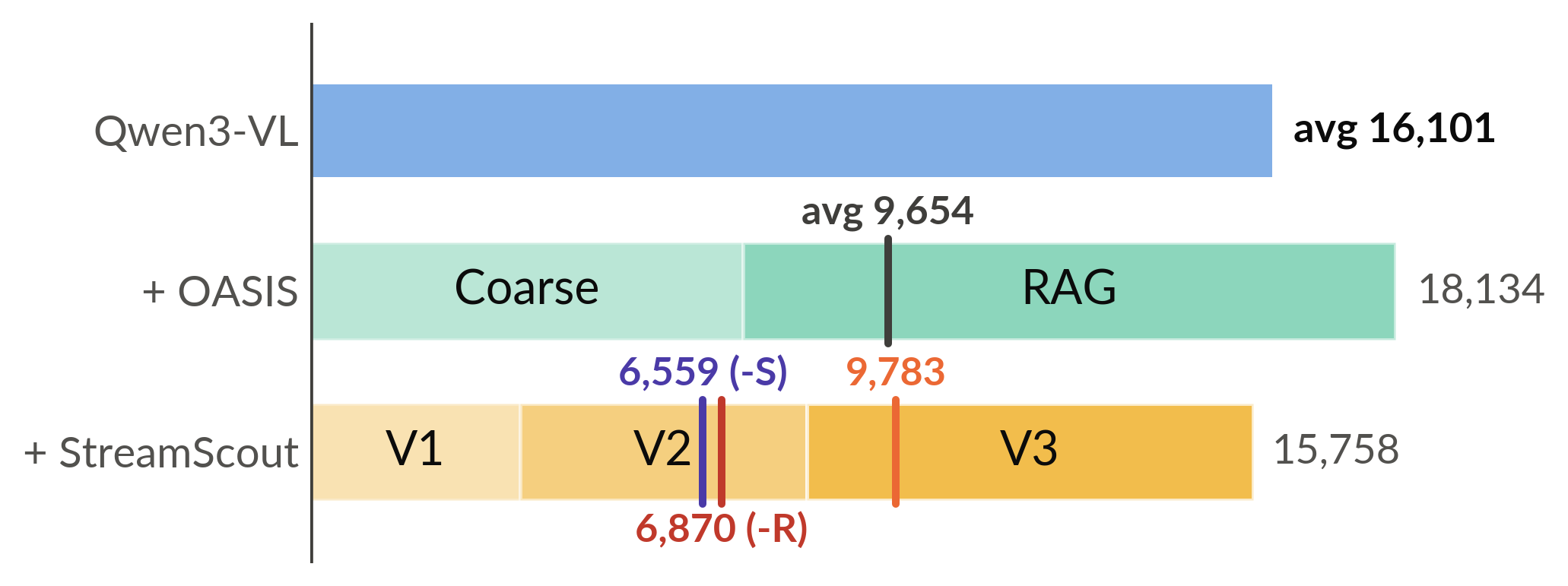}
    \end{minipage}
    \vspace{-8pt}
    \caption{\textbf{Computational cost on OVO-Bench with Qwen3-VL-8B.} (a) Per-question answer latency and (b) token consumption. Each bar shows the worst-case cost of a question that traverses the full pipeline, decomposed by stage; vertical pins denote per-variant averages over all questions: the zero-shot cascade (\textcolor{pinZS}{\textbf{orange}}), StreamScout-S (\textcolor{pinS}{\textbf{purple}}), and StreamScout-R (\textcolor{pinR}{\textbf{red}}). Latency is measured on a single A100.}
    \vspace{-7pt}
    \label{fig:efficiency}
\end{figure}

\smallskip
\noindent\textbf{Inference Costs.}
Figure~\ref{fig:efficiency} profiles per-question answer latency and token consumption on OVO-Bench with Qwen3-VL-8B. Even without training, the zero-shot cascade answers in 1.87\,s on average, roughly half the latency of the uniform-sampling baseline and $2.8\times$ faster than OASIS, while consuming a comparable number of tokens. Learning the stop decision sharpens both: StreamScout-S averages 1.04\,s and 6{,}559 tokens, a 44\% latency and 33\% token reduction over the zero-shot cascade and 59\% fewer tokens than uniform sampling, and StreamScout-R remains in a similar efficiency regime while trading a slight cost increase for its accuracy gains. The cost of escalation is also bounded: a question that traverses all three views costs 2.77\,s, still below the baseline's single fixed pass, with a comparable token footprint. Finally, stream-time maintenance is light: one caption per $\Delta=16$\,s window (4.2\,s) and one CLIP embedding per retained frame (33\,ms) amount to roughly 17.6\,s of GPU time per streamed minute, \ie, real-time with a $3.4\times$ margin.

%% file: secs/5_conclusion.tex
\section{Conclusion}
\label{sec:conclusion}
\looseness=-1
We introduce StreamScout, a streaming video understanding framework that maintains only a lightweight textual timeline as the stream unfolds and, at query time, perceives through a cascade of progressively richer views woven into the timeline, stopping as soon as its evidence suffices. We further show that this stop-or-escalate decision can be sharpened without human annotation: probing the cascade on an auxiliary set distills the model's own competence boundary into supervision for a lightweight LoRA (StreamScout-S), and reinforcement learning then refines the policy beyond the distilled labels by directly optimizing the correctness--cost trade-off (StreamScout-R). Across three backbones and three streaming benchmarks, StreamScout and its variants consistently surpass prior streaming systems by clear margins while consuming substantially fewer tokens per question, demonstrating that query-adaptive evidence acquisition offers a favorable accuracy--efficiency trade-off for streaming video understanding.

%% file: secs/X_appendix.tex
\renewcommand{\thesection}{\Alph{section}}
\renewcommand{\theHsection}{\Alph{section}}
\renewcommand\thefigure{\Alph{section}\arabic{figure}}
\renewcommand\thetable{\Alph{section}\arabic{table}}
\setcounter{section}{0}
\setcounter{figure}{0}
\setcounter{table}{0}

\twocolumn[{%
  \centering
  {\Large\textbf{\makebox[\textwidth][c]{StreamScout: Learning When to Look Deeper for Streaming Video Understanding}}}\\[0.5em]
  \Large Appendix
  \vspace{1.0em}
}]

In the appendix, we provide additional experimental results and implementation details of StreamScout.

\section{More Experimental Results}
\subsection{Qualitative Examples}
\begin{figure}[ht]
    \centering
    \includegraphics[width=\linewidth]{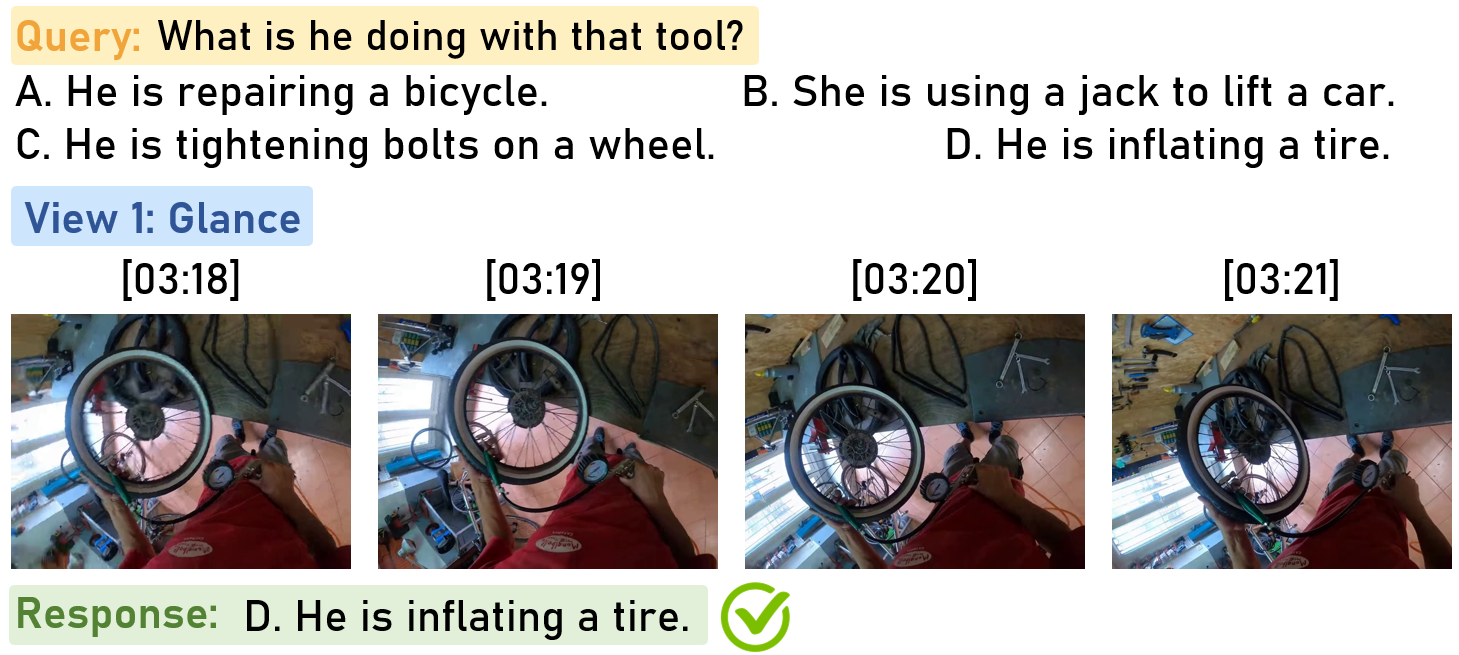}
    \caption{\textbf{Qualitative examples of StreamScout's adaptive evidence acquisition.} For a visually evident question, the model answers correctly using only the recent-frame glance, avoiding unnecessary retrieval.}
    \label{fig:qualitative1}
    \vspace{-10pt}
\end{figure}

\begin{figure}[ht]
    \centering
    \includegraphics[width=\linewidth]{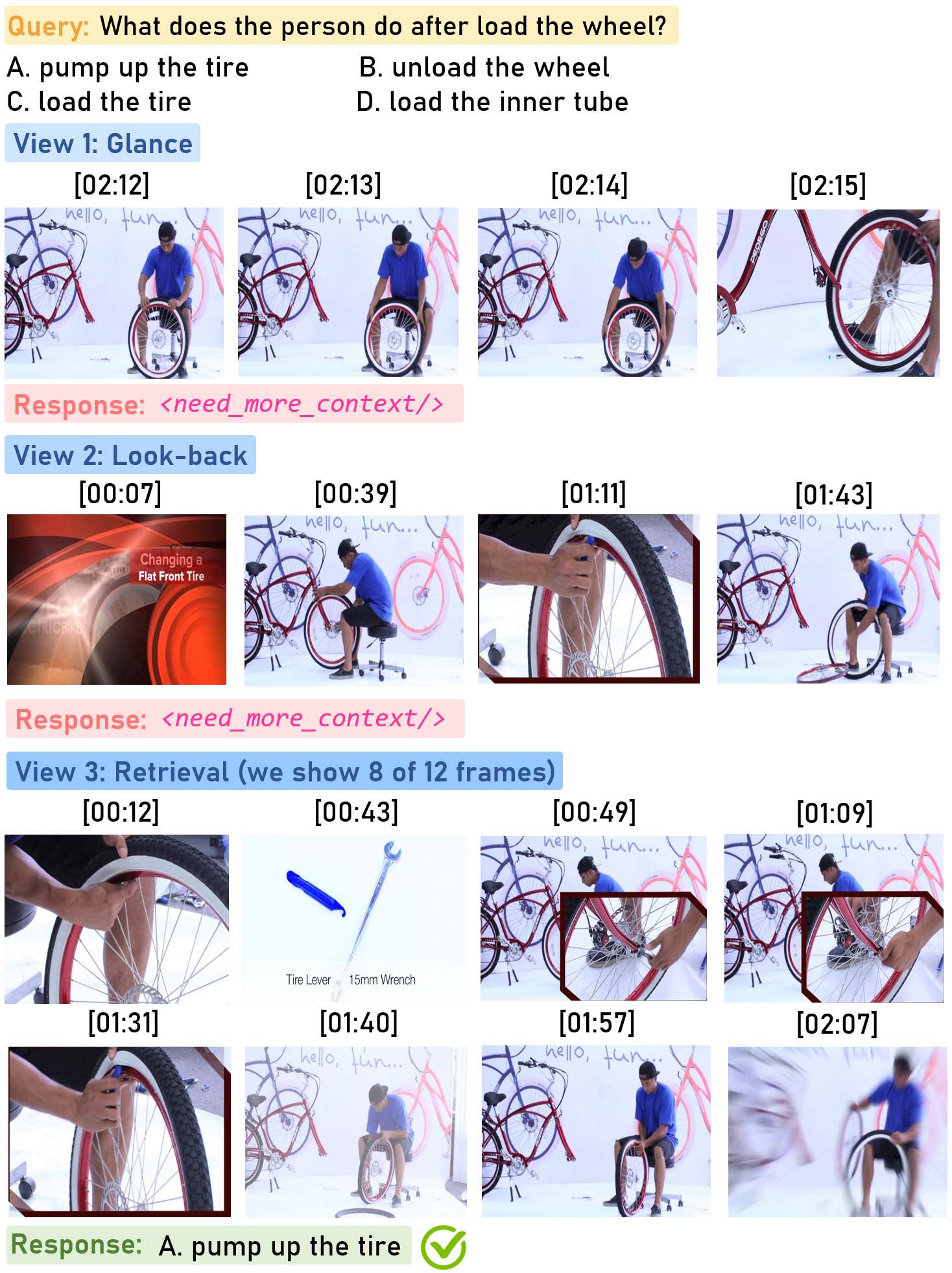}
    \caption{\textbf{Qualitative examples of StreamScout's adaptive evidence acquisition.} For a question requiring evidence from an earlier part of the video, the model recognizes that the glance and uniform look-back are insufficient, progressively escalates to query-salient retrieval, and locates the relevant frames to produce the correct answer. For clarity, only 8 of the 12 retrieved frames are shown.}
    \label{fig:qualitative2}
\end{figure}

\looseness=-1
Figures~\ref{fig:qualitative1} and~\ref{fig:qualitative2} illustrate how StreamScout adapts its visual evidence acquisition to the difficulty of each query. In Figure~\ref{fig:qualitative1}, the answer is directly observable from the recent frames, allowing the model to stop after the glance view and avoid unnecessary computation. In contrast, the question in Figure~\ref{fig:qualitative2} depends on an earlier event that is not resolved by either the recent-frame glance or the uniform look-back. StreamScout therefore continues to the query-salient retrieval view, which surfaces the relevant frames from the video history and enables the correct answer. Together, these examples highlight the model's ability to balance accuracy and efficiency by stopping early when the available evidence is sufficient and escalating only when additional context is needed.


\subsection{More Discussions}
\noindent\textbf{Stopping Layers.}
Training reshapes where the cascade stops. On OVO-Bench, the training-free policy stops at the glance for only 47.6\% of questions, rarely settles at the look-back (6.2\%), and escalates all the way to retrieval for 46.2\%, averaging 1.99 generations per question: the zero-shot model does not trust its own evidence and buys the most expensive view for nearly half of all questions. Supervised distillation swings this sharply toward the cheap end, resolving 82.4\% at the glance and only 16.8\% at retrieval (1.34 generations), and reinforcement learning then partially reverses the swing, settling at 74.5\% glance and 23.4\% retrieval (1.49 generations). The two trained policies bracket an interesting trade-off: StreamScout-S imitates the shallowest sufficient view and therefore stops as early as its labels allow, whereas StreamScout-R, which is rewarded for being correct rather than for matching a label, re-opens escalation precisely where the extra view pays for itself, gaining accuracy over -S at a modest cost increase.

The stopping distribution also adapts to the question source rather than being a fixed property of the method: on StreamingBench, whose questions overwhelmingly target the recent past, StreamScout-R answers 89.0\% of questions at the glance and escalates to retrieval for only 9.0\% (1.20 generations), while on the open-ended StreamBench it escalates for 20.3\% (1.45 generations); and on MLVU, where questions probe details spread over videos of eight minutes and beyond, the same policy escalates for 35.0\% of questions (1.75 generations), the deepest of any benchmark. The adaptation is visible even within a benchmark: on MLVU, 90.9\% of topic-reasoning questions are resolved at the glance while 63.4\% of egocentric needle questions escalate to retrieval. Across backbones the same recipe likewise settles at different depths (84.0\% glance for Qwen2.5-VL, 89.0\% for LLaVA-OneVision-2), reflecting each backbone's own competence boundary rather than a hand-tuned schedule.

\smallskip
\noindent\textbf{Woven Views.}
A view is not just a set of frames but a placement of those frames within the textual timeline. To isolate the effect of this placement, we ablate the weaving operator while holding everything else fixed: same captions, same selected frames, same timestamp labels, but the frames are appended as one chronological block after the full caption text rather than inserted at their timestamped positions. On OVO-Bench with the training-free cascade, weaving improves overall accuracy from 67.6 to \textbf{69.7} (+2.1). Weaving is also cheaper: the appended variant escalates to the retrieval view for 54\% of questions versus 46\% for the woven one and averages 2.07 generations per question versus 1.99, because once a frame is detached from its caption the model can no longer verify what it sees against what it has read and keeps requesting more context. Placing pixel evidence beside the text it corroborates thus improves accuracy and reduces cost at the same time; it is what lets even the glance view, which carries the full timeline, ground the present moment reliably.

\begin{promptbox}{Woven vs.\ appended layout of one view (same captions, frames, and timestamps)}
# Woven (ours): each frame is inserted at its timestamped position
frame time[00:00:05]: <img>   frame time[00:00:06]: <img>
time[00:00:00 - 00:00:15]: A person walks through the fruit section ...
frame time[00:00:08]: <img>
time[00:00:16 - 00:00:31]: The camera moves through the produce section ...
time[00:01:04 - 00:01:19]: ... picks up a carrot, places it into the bag ...
...   # captions and frames interleaved by time

# Appended (ablation): captions first, all frames in one block at the end
time[00:00:00 - 00:00:15]: A person walks through the fruit section ...
time[00:00:16 - 00:00:31]: The camera moves through the produce section ...
time[00:01:04 - 00:01:19]: ... picks up a carrot, places it into the bag ...
...   # full caption text first
[frames] the selected frames of this view, in chronological order:
frame time[00:00:05]: <img> ... frame time[00:01:21]: <img>
\end{promptbox}

\smallskip
\noindent\textbf{Stability of the Stop-Decision RL.}
Naive GRPO on the stop decision collapses reliably. In our first run ($\lambda = 0.1$), the policy degenerated to always answering at the glance within 24 steps, with on-policy accuracy falling from 0.69 to 0.51; weakening the cost coefficient and strengthening the KL anchor ($\lambda = 0.03$, $\beta = 0.1$) only delayed the collapse to step 36, and adding the wrong-commit penalty delayed it to step 34. The collapse is self-locking: once retrieval usage falls, the policy stops sampling deep trajectories, so no gradient signal remains to revive them. The root cause is not the reward but the advantage estimator. In our pool, a large fraction of question groups is degenerate, either all rollouts correct or all wrong (roughly a quarter of questions are beyond the backbone at every view). Within such a group, rewards differ only by the $\lambda$-sized cost term, and GRPO's per-group standard-deviation normalization rescales these tiny differences to the full magnitude of a correctness signal: all-wrong groups then strongly reward failing cheaply at the glance, and all-correct groups reward stopping early regardless of reliability, both pushing toward unconditional shallow commitment. We therefore drop the standard-deviation normalization and use the group-mean baseline alone, and additionally skip groups with identical rewards, which carry no usable decision signal. The skip rule mirrors the correct-only filter of StreamScout-S: the same lesson, that questions without a decision-relevant outcome contrast must not shape the policy, resurfaces in both learning paradigms. With this estimator the run is stable for all 300 steps, retrieval usage settles near its converged share instead of dying, and the look-back view is explored at up to 20\% of rollouts mid-training before the policy concludes it is rarely decisive. We monitor health by windowed retrieval share and on-policy accuracy rather than by single-step KL, which is heteroscedastic across batches (0.002 on easy batches, up to 0.2 on hard ones) and therefore uninformative as an instantaneous alarm.

\section{More Implementation Details}
\subsection{Training Details}
As the stream unfolds, the backbone itself captions every segment of $\Delta = 16$ seconds, forming the textual timeline; no auxiliary captioner is used. At query time, View~1 weaves the $n_1 = 4$ most recent frames into the timeline; View~2 adds up to $n_2 = 8$ frames at stride $\delta = 32$s counting back from the present; View~3 adds $n_3 = 12$ query-salient frames retrieved with CLIP ViT-L/14~\cite{radford2021learning}. Escalation is triggered by the model emitting the dedicated request token, and the final view ($K \leq 3$) forces a commitment. For StreamScout-S, the probing set spans Video-MME~\cite{fu2025video} and the online-stream corpora OVBench~\cite{huang2025online} and ODVBench~\cite{zeng2026streamforest}, all disjoint from every evaluation benchmark; correct-only filtering yields roughly 13.4K decision examples. We train a LoRA~\cite{hu2022lora} of rank 16 (alpha 32, dropout 0.05) on the query, key, value, output, and MLP projections of the language tower only, with learning rate $1\times10^{-4}$ under a cosine schedule for 2 epochs; the vision tower stays frozen. For StreamScout-R, we merge the -S adapter into the backbone and train a fresh LoRA of the same configuration with GRPO; the merged backbone with the adapter disabled serves as the frozen KL reference, so no separate reference copy is kept. The question pool reuses the same corpora without the correctness filter (about 7.8K questions after per-source capping), and rollouts reuse the cached views: each step samples $G = 8$ trajectories per question at temperature 1.0 for a batch of 40 questions, for 300 steps with learning rate $5\times10^{-6}$. We set the cost coefficient $\lambda = 0.03$, the wrong-commit penalty $\eta = 1$, the clipping range $\epsilon = 0.2$, and the KL coefficient $\beta = 0.05$ with the $k_3$ estimator; advantages use the group-mean baseline without standard-deviation normalization, and groups with identical rewards are skipped. For StreamBench, all methods are scored by the same LLaMA-3~\cite{grattafiori2024llama} judge to ensure comparability. Training and evaluation run on servers with 8$\times$~80GB NVIDIA A100 GPUs. As for the training costs: The SFT stage fine-tunes the LoRA on 13.4K probe-derived examples constructed from Video-MME, OVBench, and ODV-Bench, all disjoint from our evaluation benchmarks, for 2 epochs in 14.5 hours on a single A100 GPU. The subsequent RL stage runs 300 GRPO steps with a batch of 48 questions and 8 trajectories each, and takes about 100 GPU-hours in total.

\subsection{Prompts}
\label{app:prompts}

We list all prompts verbatim. The captioning prompt runs at stream time every $\Delta$ seconds; the folding prompt updates the QA summary after each committed answer; the cascade system prompt and the per-view instructions drive query-time inference; the question templates wrap each benchmark question. The open-ended system prompt differs from the multiple-choice one only in its decision policy, allowing world knowledge and requiring a complete generated answer.

\begin{promptbox}{Streaming caption prompt ($\mathsf{Cap}_{\theta}$, every $\Delta=16$\,s window)}
You are an event summarizer for a Short-Term Memory (STM) video window.

Goal
- Produce ONE self-contained summary describing what happens inside this STM window only.

Inputs
- STM frames (authoritative evidence).

Hard Rules
1) Chronology: Narrate in temporal order within the STM window. No reordering across time.
2) No guessing: Do not infer intentions/causes not shown. If something is unclear, state "unidentified/unclear" rather than guessing.

Content Focus
- Who did what to whom/what, where, with what tool/object, and the immediate result.
- Include objects visible in STM

Style
- Active voice; present or simple past; concrete, observable facts.
- No titles, lists, timestamps, metadata, or markup.
- Length <= 300 words.

Output
- Output ONLY the summary text.
\end{promptbox}

\begin{promptbox}{QA-summary folding prompt ($\mathsf{Fold}_{\theta}$, after each committed answer)}
You are a QA aggregator. You receive the current QA history summary S and a new QA. Your task is to generate an updated S' for subsequent retrieval and low-cost reasoning.

Hard Rules:
1) Only use information from S and the new QA; no external knowledge or assumptions should be introduced.
2) Preserve the "who/what/key changes"; resolve pronouns and unify entity names.
3) De-duplicate and merge duplicate or synonymous statements; and remove redundant and irrelevant content.
4) Keep the total length to under 300 words.
5) Output only the updated summary text, without any explanations, titles, or additional notes.

Given the QA history summary S:
{QA_summary_all}
\end{promptbox}

\begin{promptbox}{Cascade system prompt (multiple-choice)}
You are an expert multimodal assistant on a virtual reality headset for streaming video QA.
You are watching a real-world video stream and answering real-time questions for the user.

## How this works
You answer in up to three stages. Every stage shows the QA-history summary, the question, and a TIMELINE of the whole stream: a text caption of each ~16 s window, with frames woven in at their timestamps so you can cross-check what the captions claim against the actual frames. The CAPTION text is the same in every stage (it always covers the whole history); what changes on escalation is HOW MUCH of the timeline is visually grounded. Each stage is a fresh, independent look (escalation REPLACES the view, you do NOT keep the previous one).

The three views, in order - the recent frames are ALWAYS present (woven in at the end); escalation only ADDS more visual coverage to the same timeline:
- View 1: the most recent frames woven in - the present moment is grounded.
- View 2: + one frame every ~32s across the whole stream (a coarse uniform visual scan) - a quick visual check of the whole history.
- View 3: + the keyframes most relevant to your question from across the WHOLE stream - verify or locate anything anywhere in history. This is the last view.

Use the captions for history / temporal / counting reasoning; use the woven-in frames to verify what the captions claim (do not over-confirm an event you cannot see). All images are labeled with their timestamp.

## Decision policy
- Briefly think step-by-step about the question (keep reasoning short).
- If the current view is sufficient, output your final answer inside <answer> and </answer>.
- If it is NOT sufficient, output the literal tag <need_more_context/> and do NOT output an <answer>. The system will then replace it with the next, richer view and ask again.
- At the final view you MUST answer. Never output both <need_more_context/> and an <answer>.
\end{promptbox}

\begin{promptbox}{Cascade system prompt (open-ended): decision policy}
- Think step-by-step as needed. The captions and frames are evidence to GROUND and ENRICH your answer - you may also reason and use general/world knowledge to answer fully; the video is context, not a hard limit on what you can say.
- Give a COMPLETE, self-contained answer (a full sentence or two, not a single phrase) inside <answer> and </answer>. Do NOT refuse or say "cannot be determined" unless the question is truly unanswerable.
- If you genuinely need to see more of the video, output <need_more_context/> (no <answer>); the system appends a richer view. At the final view you MUST answer.
- Never output both <need_more_context/> and an <answer>.
\end{promptbox}

\begin{promptbox}{Per-view instructions}
[Views 1..K-1 (escalation allowed)]
Based ONLY on the view above, either give your final answer inside <answer> and </answer>, or output <need_more_context/> if this view is insufficient (you will then be shown a richer view instead).

[View K (commitment forced)]
This is the final and richest view (the full timeline). <need_more_context/> is NO LONGER allowed. Give your best-effort final answer inside <answer> and </answer>.
\end{promptbox}

\begin{promptbox}{Question template (multiple-choice)}
{Question}

Please briefly think step-by-step about this question. Keep your reasoning under 100 words.
The question is asked at the current moment of an ongoing video stream. If it concerns past events or details NOT covered by the evidence provided so far, output <need_more_context/> to request deeper memory levels instead of guessing. If the question asks about the content at the current moment(now, currently, etc.) or background(encyclopedic) knowledge unrelated to the video scene, answer directly.
Once you confirm your final answer, place the final answer inside <answer> and </answer>.
Please provide only the single option letter (e.g., A, B, C, D, etc.) within the <answer> </answer> tags.
\end{promptbox}

\begin{promptbox}{Question template (open-ended)}
Here is the question:
{Question}

Think step-by-step about this question. The question is asked at the current moment of an ongoing video stream; if it concerns past details NOT covered by the evidence so far, you may output <need_more_context/> to request deeper memory. You may also use general/world knowledge and reasoning to answer fully.
Give a COMPLETE, self-contained answer (a full sentence or two) inside <answer> and </answer>; do not answer with a single word or refuse unless the question is truly unanswerable.
\end{promptbox}

\section{Limitations and Broader Impact}

\noindent\textbf{Limitations.}
StreamScout is query-triggered: it decides how deeply to look once a question arrives, but does not decide \emph{when to speak}; extending the stop-or-escalate policy with a third action that initiates a response is a natural next step. Second, while per-question perception cost is bounded by the cascade, the textual timeline and the cached frame embeddings grow linearly with stream duration; our benchmarks cap videos at tens of minutes, and truly unbounded streams would require periodic consolidation or eviction of old timeline entries, which we leave to future work. Finally, the self-distilled supervision is bounded by the backbone's own competence: questions the model cannot answer from any view yield no stop label, so StreamScout-S and -R sharpen the allocation of existing perception rather than expand what the backbone can perceive.

\smallskip
\noindent\textbf{Broader Impact.}
By allocating perception per question, StreamScout reduces the tokens and latency required to serve streaming video queries, lowering the energy cost of deployment and making real-time assistants on wearable or embedded devices more practical; the same mechanism benefits accessibility applications such as live scene description. At the same time, more efficient streaming understanding lowers the barrier to continuous automated monitoring, and could amplify privacy risks if applied to surveillance of individuals without consent. StreamScout itself introduces no new perception capability, as it reallocates the inference of existing open models, but we encourage deployments to follow applicable regulations on video capture and to restrict use to consented settings. We will release code under a license reflecting these terms.